\documentclass[conference]{IEEEtran}
\IEEEoverridecommandlockouts

\usepackage{cite}
\usepackage{amsmath,amssymb,amsfonts}
\usepackage{algorithmic}
\usepackage{graphicx}
\usepackage{textcomp}
\usepackage{xcolor}
\usepackage{comment}

\newtheorem{defdef} {Definition}

\begin{document}

\title{Fast Evaluation of DNN for Past Dataset in Incremental Learning}

\author{\IEEEauthorblockN{Naoto Sato}
\IEEEauthorblockA{\textit{Research \& Development Group, } 
\textit{Hitachi, Ltd.}\\
naoto.sato.je@hitachi.com}

}

\maketitle

\begin{abstract}
During the operation of a system including a deep neural network (DNN), new input values that were not included in the training dataset are given to the DNN. In such a case, the DNN may be incrementally trained with the new input values; however, that training may reduce the accuracy of the DNN in regard to the dataset that was previously obtained and used for the past training. It is necessary to evaluate the effect of the additional training on the accuracy for the past dataset. However, evaluation by testing all the input values included in the past dataset takes time. Therefore, we propose a new method to quickly evaluate the effect on the accuracy for the past dataset. In the proposed method, the gradient of the parameter values (such as weight and bias) for the past dataset is extracted by running the DNN before the training. Then, after the training, its effect on the accuracy with respect to the past dataset is calculated from the gradient and update differences of the parameter values.
To show the usefulness of the proposed method, we present experimental results with several datasets. The results show that the proposed method can estimate the accuracy change by additional training in a constant time.
\end{abstract}

\section{Introduction}
\label{intro}
The introduction of machine-learning technologies in various industrial fields has been advancing. Among those technologies, deep neural networks (DNNs) are being popularly applied. In addition to replacing human tasks, in a number of fields, DNNs are outperforming people. DNNs are trained with a dataset, which is composed of pairs of input values and their corresponding expected output values. A part of the dataset is used for training DNNs, and the rest is used to evaluate the trained DNNs. In the training, when input values included in the training dataset are input, values of parameters such as weights and bias are adjusted so that the expected output values are more likely to be obtained. After the training is completed, the test dataset is used to measure the probability of obtaining the expected output value as expected. This probability---called accuracy---indicates the validity of the developed DNNs. However, the accuracy measured during development is only that with respect to the existing dataset retained at that time. That is to say, when input values that are not included in the existing dataset are given, the output values are not exclusively those expected; consequently, the accuracy of the DNN during operation is lower than that during its development.

When the accuracy of DNNs decreases during an operation, it is useful to apply incremental learning \cite{missing} \cite{incremental} \cite{appraisal} \cite{zhou2023deep} \cite{wang2024comprehensive} \cite{leo2024survey}, in which a DNN is trained by using a dataset newly acquired during the operation. In incremental learning, parameter values of the DNN (such as weights and biases) are adjusted to improve the accuracy concerning the newly acquired dataset. However, the result of the adjustment also affects the accuracy for the dataset used in the previous trainings. When the accuracy for the past dataset decreases, the updated DNN is not easily adopted. If a system operator finds that a concept drift of input values happens, the decrease in accuracy for the past dataset can be ignored because the past dataset and its similar data are not input thereafter. However, since the operator is not always able to detect changes in the distribution of input values, they can rarely be confident that the decrease in the accuracy for the past dataset can be ignored. Even if the distribution changes due to concept drift, a part of the input values before the change may still be included after the change. In such cases, it should maintain a certain level of accuracy for the past dataset. Therefore, the decision to use the updated DNN is based on the accuracy for the past dataset and the results of the concept drift evaluation. In addition, in a number of application systems, the decision to adopt or discard the updated DNN should be made as soon as possible to prevent losing business opportunities. To quickly determine whether to adopt the updated DNN, a quick evaluation of the accuracy for the past dataset is necessary.

To grasp the effect of additionally conducted training on the accuracy for the past dataset, it is sufficient to run the updated DNN with all the input values included in the past dataset and acquire its accuracy  again. However, when the number of input values in the past dataset is huge, it takes a long time to execute that test. In other words, the accuracy of the DNN cannot be evaluated quickly by the execution.

We propose a method for quickly evaluating the effect of additional training on the accuracy for the past dataset \cite{sato2018}. In the proposed method, the gradient of the parameter values is extracted by executing the DNN before updating the input values in the past dataset. Then, after the additional training, its effect on the accuracy for the past dataset is evaluated on the basis of the gradient and the update differences of parameter values of the DNN. We also leverage a linear regression analysis to estimate the increase/decrease in accuracy. The calculation amount to be performed after additional training does not depend on the number of input values in the past dataset. Therefore, even if the number of input values in the past dataset is huge, applying the proposed method makes it possible to evaluate the effect of additional training quickly. To demonstrate the usefulness of the proposed method, we show the experimental results using the MNIST dataset, Fashion MNIST, and The German Traffic Sign Recognition Benchmark dataset.

The rest of this paper is organized as follows. In Section \ref{prelim}, the structure of DNNs and a flow of incremental learning are defined. The problem that we focus on in incremental learning is also described. In Section \ref{method}, the proposed method is defined on the basis of several calculation formulas. In Section \ref{exp}, the results of experiments applying the proposed method to three datasets are shown. In Section \ref{eval}, the usefulness of the proposed method is evaluated and discussed on the basis of the experimental results. In Section \ref{relwork}, related work is described, and in Section \ref{conclusion}, the conclusions drawn from this study are presented.

\section{Preliminaries}\label{prelim}
\subsection{Deep Neural Networks}
For an arbitrary DNN, denoted as $N$, handling a classification problem of class $c (c>1)$, $I$ is taken as the number of neurons making up $N$, and each neuron (in any layer) is denoted as $n_i (1 \leq i \leq I)$. Also, $J_i$ is taken as the number of parameters used to calculate the value of $n_i$, and the parameter itself is denoted as $[w_{i,1},..., w_{i,j},...,w_{i,J_{i}}]$. Note that if $n_{i_1}$ and $n_{i_1}$ are included in different layers, $J_{i_1}$ and $J_{i_2}$ can be different. Then, a vector obtained by combining the parameters of all neurons is defined as $W = [w_{1,1},..., w_{1,J_{1}}, w_{2,1},..., w_{2,J_{2}},..., w_{I,1},..., w_{I,J_{I}}] $. For a multilayer perceptron, for example, these parameters correspond to weights and biases. Note that the formula for calculating the value of each neuron by using these parameters is not described in this paper.

For an arbitrary input value $x^m$, the expected corresponding output value is expressed as $t(x^m)$. $t(x^m)$ represents the identifier of the classification class to which $x^m$ belongs. When $x^m$ is input to $N$, the output value returned by $N$ is expressed as $y(x^m)$, which corresponds to the $c$-dimensional vector $[y_{1}^m, ..., y_{k}^m, ..., y_{c}^m]$. The $y_k^m$ value for each dimension represents the probability that the input value $x^m$ belongs to class $k$. If $y_k^m$ has the largest value in $[y_{1}^m, ..., y_{k}^m, ..., y_{c}^m]$, class $k$ is denoted as $fst(y(x^m))$. That is, $ \forall y_k^m \cdot y_k^m \leq y_{fst(y(x^m))}^m$ holds. When $fst(y(x^m)) = t(x^m)$ holds, it is said that $N$ can correctly classify $x^m$. Likewise, if $y_k^m$ has the second-largest value, class $k$ is denoted as $snd(y(x^m))$. Namely, $ \forall y_k^m \cdot y_k^m \neq y_{fst(y(x^m))}^m \Rightarrow y_k^m \leq y_{snd(y(x^m))}^m$ holds.

Since the number of layers, neurons, and parameters are generalized in the definition of DNNs and types of activation functions are not specified, the proposed method can be applied to any neural networks in which the gradient of parameters is computable.

\subsection{Incremental Learning}\label{additional}
\begin{figure}[!t]
\centering
\scalebox{0.40}{\includegraphics[bb=0 0 634 300]{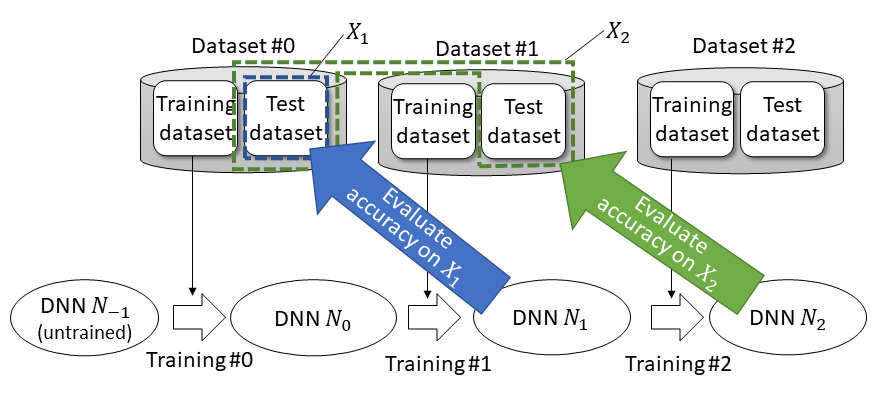}}
\caption{Flow of incremental learning}
\label{fig01}
\end{figure}

The flow of incremental learning assumed is shown in Fig. \ref{fig01}.

In this flow, an untrained DNN $ N_{-1}$ is trained first by using the training dataset of dataset \#0. The trained DNN $N_0$ is evaluated by using test dataset of dataset \#0. After testing, $N_0$ starts to be used. During its use, dataset \#1 is newly acquired. Then, $N_0$ is trained again using the training dataset of dataset \#1, and the updated DNN $N_1$ is evaluated in the same way using the corresponding test dataset. However, $N_1$ does not always maintain sufficient accuracy for dataset \#0. Accordingly, it is necessary to evaluate the effect of training \#1 on the accuracy for dataset \#0. That necessity holds for the following incremental learning.

Incremental learning can be categorized as domain-, task-, and class-incremental learning \cite{scenario}. We focus only on domain-incremental learning \cite{mirza2022efficient} \cite{wang2024multi}. Namely, we assume that the input distribution may change, but the number of classification class $c$ is not increased and no other tasks are added through the incremental learning.

As for a DNN $N_{s} (1 \leq s)$, if a dataset for which the effect should be evaluated is expressed as $X_s$, $X_s$ is defined as follows:
\begin{defdef}
\label{def01}
\begin{eqnarray*}
X_s = \begin{cases}
 TD_{0} & (s=1) \\
 X_{s-1} \cup TD_{s} & (otherwise) \\
\end{cases}
\end{eqnarray*}
\end{defdef}
where $TD_{s}$ represents the test dataset of dataset \#$s$.

\subsection{Problem Statement}
\label{problem}
Incremental learning is useful to adjust a DNN to changes in the distribution of input values, which is known as concept drift. 
However, even if the distribution of input values changes due to concept drift, it does not necessarily mean that the input values are completely changed to different values. For example, when the range of input values is expanded, the past input values are included in the new ones.
In such cases, an operator of the system using the DNN are concerned about accuracy with respect to the past dataset. In actual operations, the operator may not be able to determine if concept drift actually occurs. Therefore, if the operator finds that the accuracy for the past dataset decreases unexpectedly, the updated DNN may be rejected.
Thus, adjusting to the new dataset and maintaining performance for the past dataset are required to be partially compatible, while being a trade-off. Therefore, the decision to operate the updated DNN is carefully made on the basis of the accuracy on the newly obtained dataset, the accuracy on the past dataset, and the results of the concept drift evaluation. When the updated DNN is not adopted, either the training is retried or a rollback to the past DNNs is executed.

To evaluate the accuracy of the updated DNNs with respect to the past dataset, it is sufficient to test the DNNs with all the data in the past dataset as input. For incremental learning, the size of the past dataset gradually increases, and the time required to evaluate the change in accuracy with respect to the past dataset also increases. Then, until the test execution on the past dataset is completed, the DNN before or after the update is tentatively selected and used. If the operator does not have any information of the change in accuracy for the past dataset, the worse DNN may be selected. This results in undesirable business losses. Therefore, we propose a method to ``quickly'' estimate the change in accuracy with respect to the past dataset. It enables the operator to select a tentative DNN with consideration of the change in accuracy for the past dataset. 

\begin{figure}[!t]
\scalebox{0.45}{\includegraphics[bb=0 0 558 220]{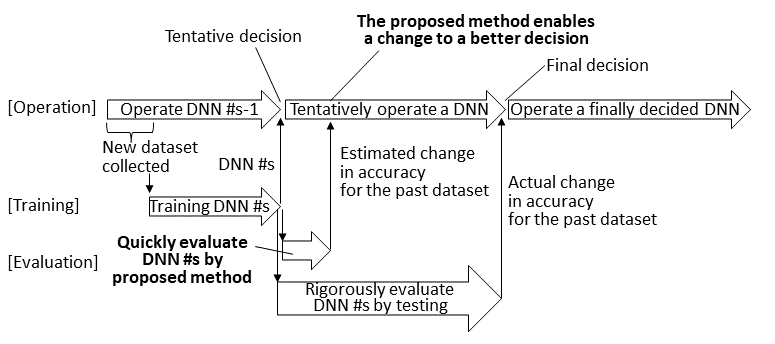}}
\caption{Usage of the proposed method}
\label{fig_add01}
\end{figure}
As shown in Fig. \ref{fig_add01}, by using the proposed method, the tentative DNN can be selected by referring to the estimated change in accuracy for the past dataset until the actual change in accuracy is obtained by testing. This could enable better DNN operation, and potentially reduce the business losses.

The proposed method is useful for systems with the following characteristics. First, the external environment of the system, i.e., the distribution of input values, gradually changes and there are both micro and macro trends of change. Secondly, the system is real-time \cite{testbed} and able to respond immediately to changes in its external environment. Examples of such systems are social infrastructure systems, such as weather prediction, power control, and so on. In the case of a system that forecasts electricity demand, even if there is a change in the past several weeks, it may be transient due to peculiar weather conditions. Therefore, the operator may be discouraged from updating the DNN if the accuracy for the past datasets is significantly decreased. Another example is a stock price forecasting system. Updating the DNN based on micro trend changes could lead to significant business losses.

\section{Proposed Method}
\label{method}
\subsection{Positive and Negative Gradients}
It is supposed that DNN $N_s$ is developed by the $s$th training on DNN $N_{s-1}$. The DNN deals with a classification problem with class $c (c>1)$. For any input value $x^m$ contained in $X_s$, the corresponding expected output value $t(x^m)$ is defined. Since the effect of the DNN update is evaluated for each classification class in the proposed method, we define $X_s^k \subseteq X_s$ as a set of input values with the same classification class $k$. The purpose of the proposed method is to estimate the change in accuracy for $X_s^k$ when $N_{s-1}$ is updated to $N_s$.

Consider the case where DNN $N_{s-1}$ fails in the inference of input value $x^m \in X_s^k$, i.e., $fst(y(x^m)) \neq t(x^m)$. In this case, $x^m$ can be a factor that improves the accuracy of $N_{s-1}$. When $N_{s-1}$ is updated to $N_s$, if $N_s$ succeeds in the inference of $x^m$, then the accuracy is increased. On the basis of this, we would like to estimate how much the likelihood of successful inference of $x^m$ increases when updating DNN $N_{s-1}$ to $N_s$. Therefore, in the proposed method, we focus on value changes of the c-dimensional vector $y(x^m) = [y_{1}^m, ... , y_{k}^m, ... , y_{c}^m]$. To avoid confusion between $y(x^m)$ in $N_{s-1}$ and $y(x^m)$ in $N_{s}$, the output value of $N_{s-1}$ is hereafter described as $y(x^m) = [y_{1}^m, . , y_{k}^m, ... , y_{c}^m]$ and the output value of $N_s$ is $y^{\prime}(x^m) = [y_{1}^{\prime m}, ... , y_{k}^{\prime m}, ... , y_{c}^{\prime m}]$.

If $N_{s-1}$ fails in the inference, it means that $y_{t(x^m)}^m$ is not the largest in $y(x^m)$, i.e. $y_{t(x^m)}^m < y_{fst(y(x^m))}^m$. Assume that updated DNN $N_s$ infers the output value of $x^m$ successfully. In that case, $y_{t(x^m)}^{\prime m}$ has increased or $y_{fst(y(x^m))}^m$ has decreased so that $y_{t(x^m)}^{\prime m} \geq y_{fst(y(x^m))}^{\prime m}$ holds. Therefore, we focus on the value change of $y_{t(x^m)}^m$ and $y_{fst(y(x^m))}^m$ to estimate the likelihood of successfully inferring the output value of $x^m$ with $N_s$. The more $y_{t(x^m)}^m$ increases, the more likely the output value of $x^m$ is successfully inferred. This is also true the more $y_{fst(y(x^m))}^m$ decreases, the more likely the output value of $x^m$ is successfully inferred. Note that $fst(y(x^m))$ is the value with respect to $N_{s-1}$. That is, $y_{fst(y(x^m))}^{\prime m}$ represents the value after $y_{fst(y(x^m))}^m$, the largest element value of $y(x^m)$, is changed by updating the DNN from $N_{s-1}$ to $N_s$.

Similarly, consider the case where accuracy is decreased by updating the DNN. When the inference result of $x^m$ is changed from success to failure, i.e., from $y_{t(x^m)}^m = y_{fst(y(x^m))}^m$ to $y_{t(x^m)}^{\prime m} \neq y_{fst(y'(x^m))}^{\prime m}$. The factors that cause the accuracy to decrease are the values of $y_{fst(y(x^m))}^m$ and $y_{fst(y'(x^m))}^{\prime m}$. Here, $fst(y'(x^m))$ denotes the class with the largest element value in $y'(x^m)$. Since the aim of this method is to estimate the effect on $X_s^k$ without running $N_s$ with $X_s^k$, it is assumed that $fst(y'(x^m))$ is not known. Therefore, $snd(y(x^m))$ is used instead of $fst(y'(x^m))$ in the proposed method. This is based on the assumption that ``If the probability corresponding to class $t(x^m)$ in $N_s$ is not the largest, then the class with the largest probability in $N_s$ is likely to be that with the second largest probability in $N_{s-1}$.'' Here, we assume that $N_{s-1}$ succeeds in the inference of $x^m$, so the class with the largest probability in $N_{s-1}$ is $t(x^m)$. If a class different from $t(x^m)$ has the largest value in $N_s$, it is natural to believe that the class is highly likely to be $snd(y(x^m))$, the class with the second highest probability in $N_{s-1}$. Therefore, to estimate the possibility that $x^m$ fails in the inference after updating the DNN, we focus on changes in $y_{t(x^m)}^m$ and $y_{snd(y(x^m))}^m$. The more $y_{t(x^m)}^m$ decreases, the more likely it is that the inference of $x^m$ fails. This is also true the more $y_{snd(y(x^m))}^m$ increases, the more likely the inference of $x^m$ fails.

In the proposed method, $X_s^k$ is input into $N_{s-1}$ and the output values are inferred before updating from $N_{s-1}$ to $N_{s}$. The set of input values that fail or succeed in the inference are $X_{F} \subseteq X_s^k$ and $X_{T} \subseteq X_s^k$, respectively. Here, $X_{F} \cap X_{T} = \emptyset $. For the sake of simplicity, $X_{T}$ and $X_{F}$ are not given $s$ and $k$ as subscripts. For $x^m \in X_{F}$, $fst(y(x^m)) \neq t(x^m)$ holds. Similarly, for $x^m \in X_{T}$, $fst(y(x^m)) = t(x^m)$ holds.

For $x^m \in X_{F}$, we define the {\it positive loss} $PL$ to evaluate the change in value of $y_{t(x^m)}^m$ and $y_{fst(y(x^m))}^m$, which cause an accuracy increase. $L$ is an arbitrary loss function. The parameters of $L$ are the output value $y(x^m)$ and the supervisory signal. Updating the DNN so that $PL$ decreases increases the likelihood of a successful inference of $x^m \in X_{F}$. For $x^m \in X_{T}$, we evaluate the value change of $y_{t(x^m)}^m$ and $y_{snd(y(x^m))}^m$, which cause accuracy decrease. Then, we define the {\it negative loss} $NL$. If the DNN is updated so that $NL$ decreases, it is more likely to fail in the inference of $x^m \in X_{F}$.

\begin{defdef}
\label{def03}
\begin{eqnarray*}
PL(x^m) &=& L(y(x^m), t(x^m)) - L(y(x^m), fst(y(x^m))) \\
NL(x^m) &=& L(y(x^m), snd(y(x^m))) - L(y(x^m), t(x^m))
\end{eqnarray*}
\end{defdef}

Next, the gradient of parameter $W_{s-1} = [w_{1,1},..., w_{i,j},...,w_{I,J_{I}}]$ of $N_{s-1}$ with respect to $PL(x^m)$ is calculated. The gradient of $PL(x^m)$, called the {\it positive gradient}, is expressed by $ \nabla PL(x^m)$. Similarly, the gradient of $NL(x^m)$, called the {\it negative gradient}, is expressed by $ \nabla NL(x^m)$. $ \nabla PL(x^m)$ and $ \nabla NL(x^m)$ are defined as follows:

\begin{defdef}
\label{def04}
\begin{eqnarray*}
\nabla PL(x^m) = \left[ \frac{\partial PL(x^m)}{\partial w_{1,1}}, ..., \frac{\partial PL(x^m)}{\partial w_{i,j}}, ..., \frac{\partial PL(x^m)}{\partial w_{I,J_{I}}} \right] \\
\nabla NL(x^m) = \left[ \frac{\partial NL(x^m)}{\partial w_{1,1}}, ..., \frac{\partial NL(x^m)}{\partial w_{i,j}}, ..., \frac{\partial NL(x^m)}{\partial w_{I,J_{I}}} \right].
\end{eqnarray*}
\end{defdef}

\subsection{Effect Estimation}
\label{influence}
After $ \nabla PL(x^m)$ and $ \nabla NL(x^m)$ are created for $N_{s-1}$, training \#$s$ is additionally executed, and DNN $N_s$ is created. Parameter $W_s$ of $N_s$ is compared with parameter $W_{s-1}$ of $N_{s-1}$, and the update difference of parameter $ \Delta W_s = W_s - W_{s-1}$, is acquired. By using $ \Delta W_s$, positive effect $PI(x^m, N_s)$ and negative effect $NI(x^m, N_s)$ to $x^m$ is calculated as follows:

\begin{defdef}
\label{def05}
\begin{eqnarray*}
PI(x^m, N_s) &=& - \nabla PL(x^m) \cdot \Delta W_s \\
NI(x^m, N_s) &=& - \nabla NL(x^m) \cdot \Delta W_s.
\end{eqnarray*}
\end{defdef}

\begin{figure}[!t]
\centerline{\scalebox{0.50}{\includegraphics[bb=0 0 520 338]{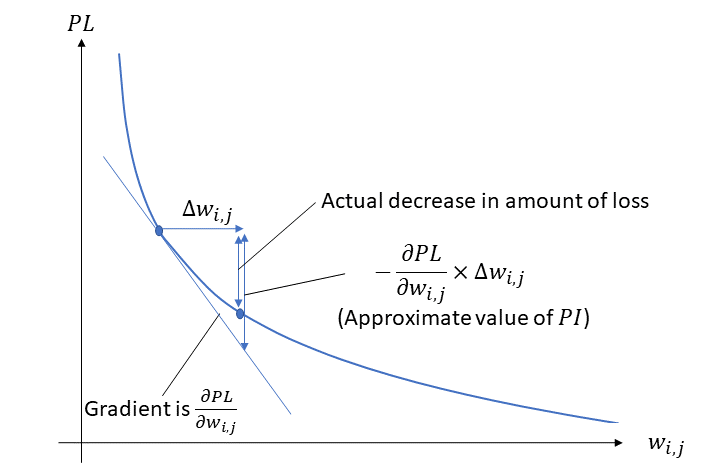}}}
\caption{Decrease in $PL$ and its approximate value}
\label{fig02}
\end{figure}

$PI(x^m, N_s)$ approximates the amount by which $PL(x^m)$ is decreased by updating parameter $W_{s-1}$ to $W_s$. Likewise, $NI(x^m, N_s)$ corresponds to the approximate decrease value in $NL(x^m)$. If $\Delta W_s$ is denoted as $[\Delta w_{1,1}, ..., \Delta w_{i,j},..., \Delta w_{I,J_{I}}]$, $PI(x^m, N_s)$ corresponds to $ \sum_{i, j} \left( - \frac{\partial PL(x^m)}{\partial w_{i,j}} \times \Delta w_{i,j} \right)$. For example, in the case that loss function $L$ is cross entropy, the relation between the decrease value in $PL(x^m)$ due to updating $w_{i,j}$ by $ \Delta w_{i,j}$, and its approximate value $- \frac{\partial PL(x^m)}{\partial w_{i,j}} \times \Delta w_{i,j}$ is shown in Fig. \ref{fig02}.

The same loss function is used as $L$ as one used for training $N_{s-1}$ from $N_{s-2}$. The training aims to reduce the loss calculated on the basis of $L$ (hereafter, training loss), in which the expected output value is the supervisory signal of the training. In this case, the output value of $N_s$ in regard to $x^m$ is more likely correct. This means that the decrease value in training loss is a barometer for evaluating the change in accuracy. It can therefore be assumed that $PI(x^m, N_1)$, which is an approximate decrease value in $PL(x^m)$, can be taken as a barometer for evaluating the change of the output value with respect to $x^m$. This is also true for $NL(x^m)$.

The decrease value in $PL(x_m)$ indicates the extent to which the likelihood of successfully inferring the output value of $x^m \in X_{F}$ is increased. Therefore, the sum of $PL(x_m)$ is the barometer of the accuracy increase in $X_{F}$. Similarly, the decrease value in $NL(x_m)$ indicates the extent to which the likelihood of failing in inferring the correct output value of $x^m \in X_{T}$ is increased. Thus, the sum of $NL(x_m)$ is a barometer of the accuracy decrease in $X_{T}$. On the basis of the aforementioned considerations, the following $EF(X_s^k, N_s)$ can be used as a barometer for evaluating the change in accuracy with respect to $X_s^k = X_{F} \cup X_{T}$.

\begin{defdef}
\label{def06}
\begin{eqnarray*}
EF(X_s^k, N_s) = \sum_{x^m}^{X_{F}} PI(x^m, N_s) - \sum_{x^m}^{X_{T}} NI(x^m, N_s)
\end{eqnarray*}
\end{defdef}

The purpose of the proposed method is to quickly estimate the change in accuracy for the past dataset immediately after updating the DNN. To achieve this, calculations that can be performed before updating the DNN should be performed in advance, and the amount of calculation after updating the DNN should be minimized as much as possible. However, only formulas of Definitions \ref{def03} to \ref{def04} can be performed before the DNN is updated. Since parameter $ \Delta W_s$ appears in Definition \ref{def05}, subsequent calculations can only be executed after updating the DNN. Moreover, since the computational complexity of the formulas in Definitions \ref{def05} and \ref{def06} depends on the number of input values contained in $X_s^k$, the amount of calculations performed after updating the DNN will increase as incremental learning progresses. Therefore, the calculations defined by Definitions \ref{def05} and \ref{def06} are replaced with the following ones defined in Definitions \ref{def07} and \ref{def08}.

\begin{defdef}
\label{def07}
\begin{eqnarray*}
GradSum(X_s^k) = \sum_{x^m}^{X_{F}} (- \nabla PL(x^m)) - \sum_{x^m}^{X_{T}} (- \nabla NL(x^m))
\end{eqnarray*}
\end{defdef}

\begin{defdef}
\label{def08}
\begin{eqnarray*}
EF(X_s^k, N_s) = GradSum(X_s^k) \cdot \Delta W_s
\end{eqnarray*}
\end{defdef}

The formula in Definition \ref{def07} can be executed before the DNN is updated. On the other hand, the formula in Definition \ref{def08} can only be executed after the update. However, its computation time is constant and independent of the number of input values contained in $X_s^k$.

\subsection{Regression model}\label{regression}
$EF$ should be a barometer to evaluate changes in accuracy. More specifically, $EF$ and accuracy are expected to have a linear relationship, which means that $EF$ increases and decreases along with accuracy. This relationship is expected regardless of the dataset. However, the scale of $EF$ values and the scale of accuracy values are different for each dataset. In other words,  these scales depend on dataset type (i.e., the problem solved by the DNN) and DNN structure. Thus, there is no general way to estimate the change in accuracy from $EF$. However, if we target a specific dataset and DNN, we are able to derive a formula regressionally from the actual calculation results of the $EF$ and the change in accuracy. Therefore, the proposed method creates a linear regression model to estimate the increase/decrease in accuracy from $EF$.

First, calculate $EF$ for $X_s^k$ from the $s-1$th to $s$th training, for $N_0$ to $N_{s-1}$, respectively. That is, compute $EF(X_s^k, N_i) (1 \leq i \leq s-1)$. Also, by performing the inference of $N_0$ to $N_{s-1}$ with $X_s^k$, respectively, the actual increase or decrease in accuracy with respect to $X_s^k$ is measured. Using these $EF$ values and the increase/decrease in accuracy as inputs, a linear regression model is created. In addition, of the calculations to obtain $EF(X_s^k, N_s)$, those in Definitions \ref{def03}, \ref{def04}, and \ref{def07} are performed before training \#$s$. Then, after training \#$s$, $EF(X_s^k, N_s)$ is obtained in accordance with Definition \ref{def08}. $EF(X_s^k, N_s)$ is input to the linear regression model to estimate the value of increase or decrease in accuracy with respect to $X_s^k$ due to the update from $N_{s-1}$ to $N_s$.

The number of samples to create the linear regression model depends on the number of trainings $s$. In the experiment shown in Section \ref{exp}, When the number of training is 100 ($s=100$), the regression model is created using the results of 99 past trainings from \#1 to \#99. When creating the regression model, the interquartile range is calculated and outliers are removed from the samples. In more detail, let $IQR$ denote the interquartile range, $Q_1$ denote the lower quartile, and $Q_3$ denote the upper quartile, then samples smaller than $Q_1 - IQR * 1.5$ and larger than $Q_3 + IQR * 1.5$ are removed as outliers. The samples larger than $Q_1 - IQR * 1.5$, and $Q_3 + IQR * 1.5$ are also removed.

The amount of calculation of Definition \ref{def08} depends on the number of elements of $W_s$ (not the number of input values contained in $X_s^k$). Accordingly, even if the size of $X_s^k$ is enormous, a change in accuracy can be evaluated in a short time. For a multilayer perceptron, the number of elements of $W_s$ is given as $O(I^2)$ for the number of neurons $I$ constituting $N_s$. The computational complexity of inference by the linear regression model is independent of $X_s^k$ and $W_s$. Therefore, the calculation order of the proposed method after training \#$s$ is given as $O(I^2)$.

\subsection{Mini-batch for PL and NL}
\label{mini}
The proposed method assumes that there is sufficient time between training \#$s-1$ and $s$. That is, there is sufficient time to calculate $EF(X_s^k, N_i) (1 \leq i \leq s-1)$, to execute DNNs $N_1$ to $N_{s-1}$ with $X_s^k$ to measure changes in accuracy, and to execute formulas in Definitions \ref{def03}, \ref{def04}, and \ref{def07}. However, depending on the number of input values contained in $X_s^k$, these calculation resources may not be sufficient. In particular, since the formulas in Definition \ref{def04} calculate the gradient for each input value $x^m$, they require much time if $X_s^k$ is enormous. In such a case, $PL$ and $NL$ are calculated for each mini-batch as follows:

\begin{defdef}
\label{def09}
\begin{multline*}
PL(X_{F}^{mb}) \\
= \frac{1}{|X_{F}^{mb}|} \sum_{x^m}^{X_{F}^{mb}}( L(y(x^m), t(x^m)) - L(y(x^m), fst(y(x^m))) )
\end{multline*}
\begin{multline*}
NL(X_{T}^{mb}) \\
= \frac{1}{|X_{T}^{mb}|} \sum_{x^m}^{X_{T}^{mb}}( L(y(x^m), snd(y(x^m))) - L(y(x^m), t(x^m)) )
\end{multline*}
\end{defdef}

where $X_{F}^{mb}$ and $X_{T}^{mb}$ represents a mini-batch, and $X_{F}^{mb} \subseteq X_{F}$ and $X_{T}^{mb} \subseteq X_{T}$ hold. $|X_{F}^{mb}|$ and $|X_{T}^{mb}|$ denote the number of data contained in $X_{F}^{mb}$ and $X_{T}^{mb}$, respectively.

$PI(X_{F}^{mb})$ and $NI(X_{T}^{mb}, N_s)$ are calculated from the gradient of $PL(X_{F}^{mb})$ and $NL(X_{T}^{mb})$, and $ \Delta W_s$, respectively.

\begin{defdef}
\label{def10}
\begin{eqnarray*}
\nabla PL(X_{F}^{mb}) = \left[ \frac{\partial PL(x^m)}{\partial w_{1,1}}, ..., \frac{\partial PL(x^m)}{\partial w_{i,j}}, ..., \frac{\partial PL(x^m)}{\partial w_{I,J_{I}}} \right] \\
\nabla NL(X_{T}^{mb}) = \left[ \frac{\partial NL(x^m)}{\partial w_{1,1}}, ..., \frac{\partial NL(x^m)}{\partial w_{i,j}}, ..., \frac{\partial NL(x^m)}{\partial w_{I,J_{I}}} \right] \\
\end{eqnarray*}
\end{defdef}

\begin{defdef}
\label{def11}
\begin{eqnarray*}
PI(X_{F}^{mb}, N_s) &=& - \nabla PL(x^m) \cdot \Delta W_s \\
NI(X_{T}^{mb}, N_s) &=& - \nabla NL(x^m) \cdot \Delta W_s
\end{eqnarray*}
\end{defdef}

The larger the size of mini-batch $X_{F}^{mb}$, the more its inference results change from failure to success for more input values. As a result, accuracy is likely to increase. Similarly, if the size of mini-batch $X_{T}^{mb}$ is large, then there are more input values that may change from success to failure, and accuracy is likely to decrease. Therefore, we calculate $EF$ by multiplying $PI$ by $|X_{F}^{mb}|$ and $NI$ by $|X_{T}^{mb}|$ as described in Definition \ref{def12}.

\begin{defdef}
\label{def12}
\begin{multline*}
EF(X_s^k, N_s) \\
= (\sum_{X_{F}^{mb}}^{X_{F}} PI(X_{F}^{mb}, N_s) * |X_{F}^{mb}|) - (\sum_{X_{T}^{mb}}^{X_{T}} NI(X_{T}^{mb}, N_s) * |X_{T}^{mb}|)
\end{multline*}
\end{defdef}

Similar to the formulas of Definition \ref{def05}, $ \Delta W_s$ appears in Definition \ref{def11}. It can only be executed after the DNN is updated. Since the computational complexities of Definitions \ref{def11} and \ref{def12} depend on the size of $X_s^k$, the amount of calculation after the DNN update will increase as incremental learning proceeds. Therefore, just as the formulas of Definitions \ref{def05} and \ref{def06} were rewritten as Definitions \ref{def07} and \ref{def08}, respectively, the formulas in Definitions \ref{def11} and \ref{def12} were rewritten as follows:

\begin{defdef}
\label{def13}
\begin{multline*}
GradSum^{mb}(X_s^k) \\
= \sum_{X_{F}^{mb}}^{X_{F}} (- \nabla PL(X_{F}^{mb})) - \sum_{X_{T}^{mb}}^{X_{T}} (- \nabla NL(X_{T}^{mb}))
\end{multline*}
\end{defdef}

\begin{defdef}
\label{def14}
\begin{eqnarray*}
EF(X_s^k, N_s) = GradSum^{mb}(X_s^k) \cdot \Delta W_s.
\end{eqnarray*}
\end{defdef}

\section{Experiment}
\label{exp}
We experimentally applied the proposed method to the MNIST \cite{mnist}, Fashion MNIST \cite{fmnist}, and German Traffic Sign Recognition Benchmark (GTSRB) \cite{GTSRB} datasets. In the following sections, arguments of the functions clear from the context have been ommited.

\subsection{Setup}
\label{setup}
In this experiment, it is assumed that trainings \#0 to \#100 are conducted. Two-thirds of the total dataset is used as Dataset \#0 (see Fig. \ref{fig01}). The remaining is divided equally to create Datasets \#1 through \#100. For example, for the MNIST dataset, which contains 70,000 image data, 46,666 data is designated as Dataset \#0, and Dataset \#1 through \#100 each consist of approximately 233 data. The ratio of dividing each dataset into the training and test datasets should be the same as the ratio of the training to test datasets in the original dataset. For the MNIST dataset, 60,000 and 10,000 data are provided for the training and testing datasets, respectively, resulting in a split ratio of 6:1.

As for the DNN models, we use a multilayer perceptron with one hidden layer (composed of 1,000 neurons) in addition to the input and output layers and a CNN with two convolutional layers with a kernel size of 3x3 and stride size of 1. As for the GTSRB dataset, we use the mini-batch method described in Section \ref{mini} because the size of the images (color channels) is too large to calculate the gradient for every input value in accordance with the formulas in Definition \ref{def04}. In this experiment, the size of the mini-batch is set to 50.

In training $s=100$, $EF(X_{100}^k, N_i) (1 \leq i \leq 99)$ are calculated. In addition, the actual increase or decrease in accuracy for $N_1$ to $N_{99}$ can be calculated by inputting $X_{100}^k$ to them . Using these data as samples, a linear regression model is created. Then, by inputting $EF(X_{100}^k, N_{100})$ into this linear regression model, the change in accuracy at training \#100 is estimated. Therefore, to evaluate the proposed method, we would like to  calculate the coefficient of determination for the created regression model with test data. However, the test data to evaluate the created regression model cannot be sufficiently obtained since only the data to evaluate the regression model is that at training \#100. Even if training \#101 is performed subsequently, the target dataset $X_{100}^k$ will be updated to $X_{101}^k$, and thus a different regression model will be created at training \#101. This means that only one data is used as input for each regression model. We therefore calculate the $R^2$ score of the regression model at training \#100 using the data of training \#1 to \#99 that are used to create the regression model. If the $R^2$ score is high, the regression estimation performance for the data from training \#1 to \#99 is also high. Furthermore, we can say that the estimation performance for the data of training \#100 should be high since it is calculated in the same way as that from training \#1 to \#99.

In each training, cross entropy is used as the loss function $L$. The experiment was performed on an Ubuntu 20.04.4 LTS machine equipped with two Intel\textregistered \ Xeon\textregistered \ Gold 6132 2.6-GHz processors with 14 cores, 786-GB memory. It also has eight NVIDIA\textregistered \ Tesla\textregistered V100 NVLink GPUs.

\subsection{Results}
\begin{figure}[h]
\centerline{\scalebox{0.46}{\includegraphics[bb=0 0 569 1100]{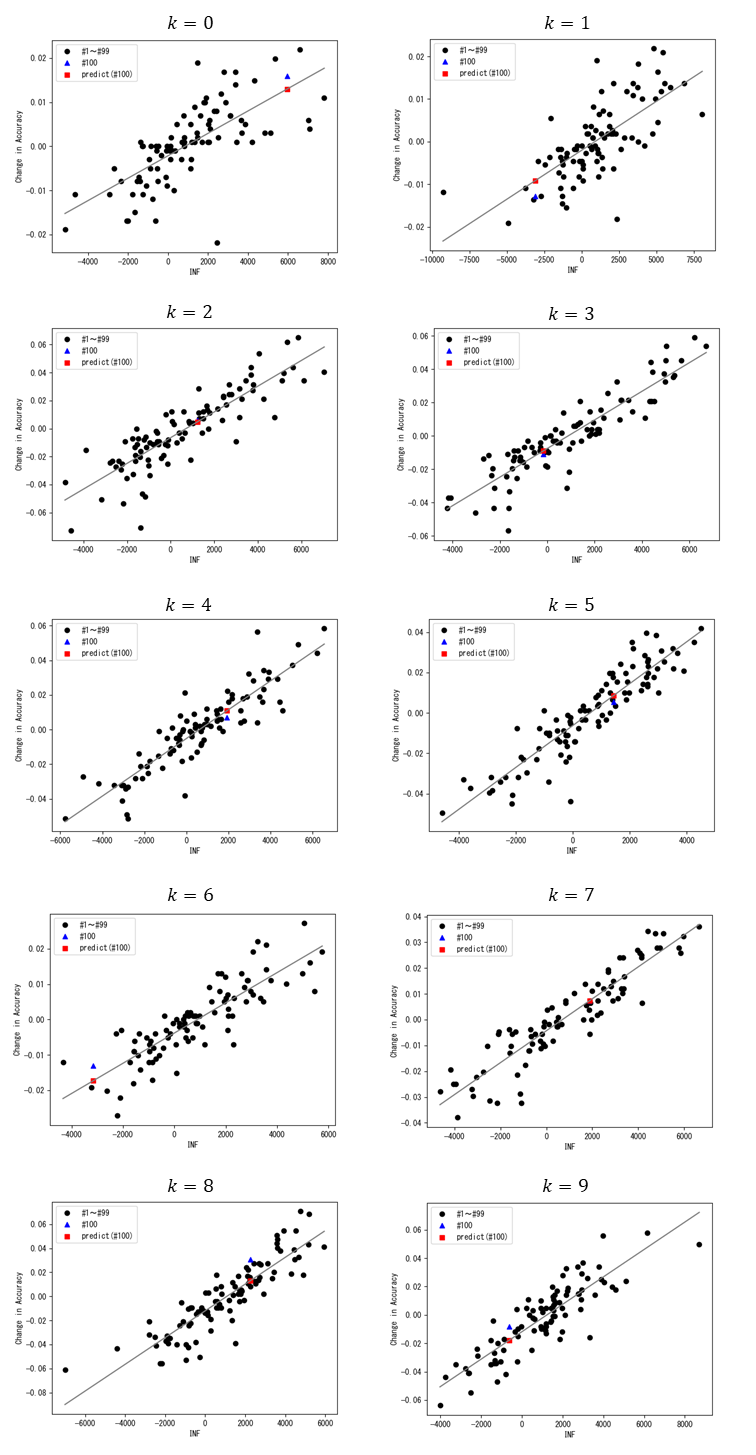}}}
\caption{Regression models of MNIST dataset}
\label{fig_add02}
\end{figure}

For the MNIST dataset, the linear regression model resulting from the experiment is shown in Fig. \ref{fig_add02}. The x- and y-axes show the values of $EF$ and the change in accuracy between before and after the training, respectively.

The dots in Fig. \ref{fig_add02} represent the data from training \#1 to \#99 used to create the regression model. However, outliers excluded on the basis of the interquartile range are not shown. The triangles represent the accuracy change estimated by the proposed method in training \#100. The rectangles represent the actual accuracy change in training \#100 calculated by executing the DNN with all input values included in $X_{100}^k$. The results for the other datasets were generally similar with a few exceptions. Cases where the results were not as expected are discussed in Section \label{eval}.

\begin{figure*}[!t]
\centerline{\scalebox{0.4}{\includegraphics[bb=0 0 1130 300]{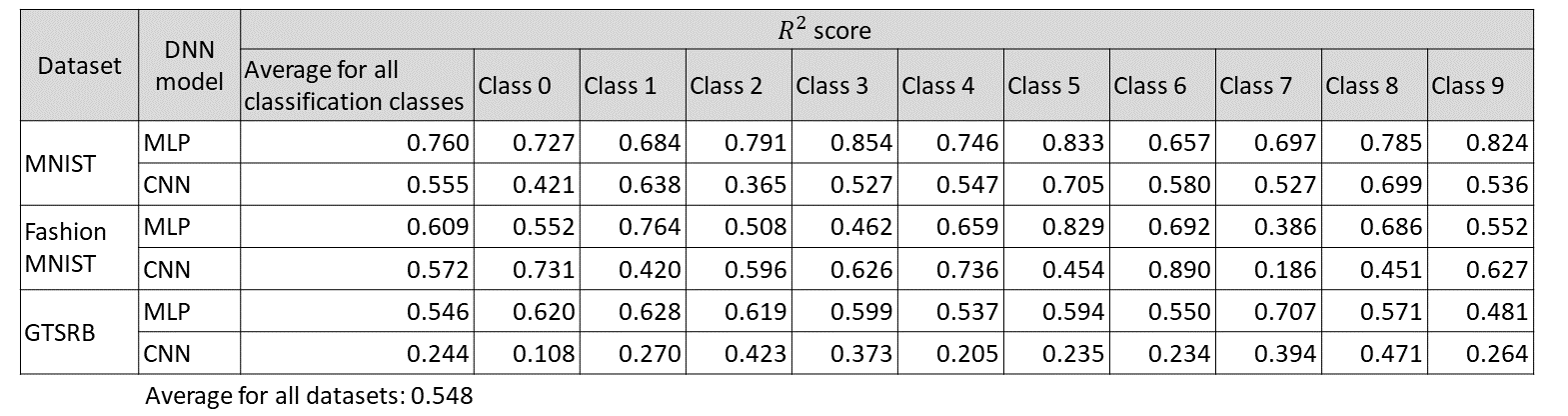}}}
\caption{$R^2$ scores of regression models}
\label{tab01}
\end{figure*}

The $R^2$ scores of the linear regression models created for the MNIST, Fashion MNIST, and GTSRB datasets are shown in Table \ref{tab01}. Only $R^2$ scores for classification classes 0 through 9 are shown. For the GTSRB dataset, the column ``Average for all classification classes'' shows the average of all 43 classes.

\begin{figure}[!t]
\centerline{\scalebox{0.60}{\includegraphics[bb=0 0 370 278]{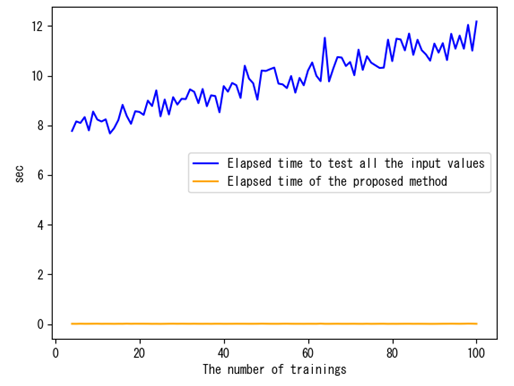}}}
\caption{Calculation times}
\label{fig03}
\end{figure}

The proposed method can be applied after training \#$s=3$ because a minimum of two samples are required to create a linear regression model. For example, in training \#$s=3$, $EF(X_{3}^k, N_1)$, $EF(X_{3}^k, N_2)$ and these corresponding changes in accuracy are used to create a regression model. For the MNIST dataset, the time taken to apply the proposed method and that to run the multilayer perceptron with all input values contained in $X_s$ for $s=3, 4, ..., 100$ are plotted in Fig. \ref{fig03}.

Fig. \ref{fig03} indicates that the test execution time increases with the number of trainings since the number of input values in $X_s$ increases. In contrast, in the proposed method, it is confirmed that the evaluation can be performed in almost the same time regardless of the increase in the number of input values. Similar results were obtained for the other datasets. Note that the time of the proposed method includes only the calculation executed after training \#$s$, that is, that of the formula in Definition \ref{def08} and the inference by the regression model.

As shown in Fig. \ref{fig03}, the proposed method does not depend on the number of input values in the past dataset. The larger the past dataset is, the more useful the proposed method is to obtain reference information for tentatively selecting a DNN until the accurate change in accuracy is confirmed by test execution. It is evident that the calculation order of the formula in Definition \ref{def08} is $O(I^2)$ for each classification class as mentioned in Section \ref{regression}. The amount of calculation corresponding to the formulas in Definitions \ref{def03}, \ref{def04}, and \ref{def07} increases in accordance with the size of the past dataset. However, those calculations are carried out before training \#$s$ and are not included in the execution time after the training. This means that the proposed method can estimate the change in accuracy in a constant time, even if the number of data included in the past dataset is enormous. However, the proposed method is useful only when the past dataset is not small. For example, in Fig. \ref{fig03}, about 10,000 input values of the MNIST dataset are inferred in training \#100, which takes only about 11 sec. If the proposed method is used for tentatively selecting DNNs as described in Section \ref{problem}, the shorter the time until the accurate change in accuracy is confirmed by testing, the less useful the proposed method will be. Assuming that the proposed method is effective if it takes more than one hour to confirm the accurate change in accuracy, the past dataset should contain more than 3 million input values, as calculated from the result of Fig. \ref{fig03}.

\section{Evaluation and Discussion}\label{eval}

\subsection{Validity of the evaluation}
As mentioned in Section \ref{setup}, the $R^2$ score shown in Fig. \ref{tab01} is calculated using the data used to create the regression model. $X_s^k$ is updated each time the number of trainings increases. Hence, in the proposed method, the regression model is re-created for each training. That is, in training \#$s$, the regression model is only used to estimate the change in accuracy for $X_s^k$ when the DNN is updated from $N_{s-1}$ to $N_s$, and is not used thereafter. In other words, the only data that can be used to evaluate the regression model is the data of training \#$s$, but one data is not sufficient for evaluation. Therefore, the $R^2$ score calculated from the data of training \#1 to \#$s-1$ is used to evaluate the regression models alternatively, These data are calculated from a DNN with the same structure and dataset $X_s^k$ as the data of training \#$s$. Therefore, it can be used to evaluate the performance of the regression model. 
For example, suppose that the residuals between the data of training \#$t (1 \leq t \leq s-1)$ and the regression model are small. This means that the regression model can accurately estimate the effect of the DNN updates in training \#$t$ for the dataset $X_s^k$. If the same is generally true for any $t$, then it is highly likely that the same regression model can accurately estimate the effect of DNN updates in training \#$s$ for the same dataset $X_s^k$. On the basis of the aforementioned consideration, we used the data of training \#1 to \#$s-1$ for the evaluation.

\subsection{Performance}
We were able to confirm that the average $R^2$ score is around 0.6, which indicates that the proposed method can be expected to perform at a certain level. In particular, if the proposed method is used for the purpose of providing reference information for selecting a tentative DNN, as described in Section \ref{problem}, using the proposed method is preferable. The results of the experiment also show that the estimation by the proposed method is not always accurate. As shown in Fig. \ref{tab01}, the $R^2$ scores of the regression models can be around 0.1 if it is low. The limitations of the proposed method should be taken into consideration when making use of it.

\begin{figure}[!t]
\centerline{\scalebox{0.55}{\includegraphics[bb=0 0 471 193]{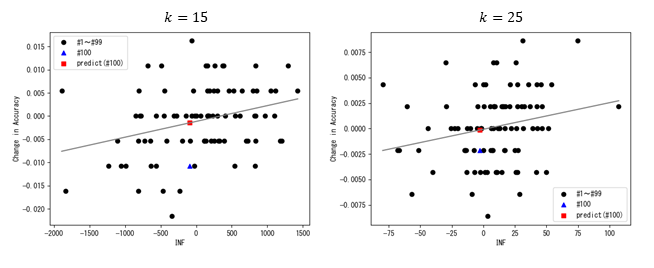}}}
\caption{Regression models for classification class $k=15$ and $25$ of GTSRB dataset}
\label{fig_add03}
\end{figure}

In a number of cases, especially for the CNN with the GTSRB dataset, the linear relationship between $EF$ and accuracy change could not be confirmed. As examples, linear regression models for the GTSRB dataset with classification class $k=15$ and $25$ are shown in Fig. \ref{fig_add03}.

In these cases, it is difficult to estimate the accuracy change by the proposed method. One common point among these cases is that the accuracy changes on the y-axis are smaller than those in Fig. \ref{fig03}. This means that the inference result may not change for many input values. In this case, the correlation between $EF$ and the change in accuracy should be low.

When DNN $N_{s-1}$ before being updated succeeds in the inference (i.e., $fst(y(x^m)) = t(x^m)$) for many $x^m in X_s^k$ but there is a large difference between the values of $y_{fst(y(x^m ))}^m$ and $y_{snd(y(x^m))}^m$, the inference results do not change even though $PI$ and $NI$ change. In this case, since the number of input values in $X_{F}$ is small, the sum of $PI$ will slightly change. For the same reason, it is also unlikely that accuracy will increase. However, since the number of input values contained in $X_{T}$ is large, the sum of $NI$ is likely to increase significantly. According to the definition of $NL$ (in Definition \ref{def03}), when $y_{fst(y(x^m))}^m$ decreases or $y_{snd(y(x^m))}^m$ increases, $NI$ increases. If there is a large difference between those values, the value of $y_{fst(y(x^m))}^m$ is seldom smaller than that of $y_{snd(y(x^m))}^m$. That is, even if the value of $NI$ increases, the relationship $y_{fst(y(x^m))}^m > y_{snd(y(x^m))}^m$ is likely to remain true and then accuracy will not decrease. Thus, the closer the value of $y(x^m)$ is to $t(x^m)$ (one-hot representation of $t(x^m)$ in reality) for most input values, the smaller the change in accuracy becomes. This determines whether the proposed method can accurately estimate change in accuracy.

\subsection{Application to each classification class}

The proposed method is applied to each classification class $k$. We realized from experiments other than ones shown in Section \ref{exp} that when the proposed method is applied to $X_s$, the $R^2$ score of the regression model is lower than in the cases of $X_s^k$. In the proposed method, the loss functions $PL$ and $NL$ are defined on the basis of the probability values of $y(x^m)$ that change the inference result. Then, the gradients of $PL$ and $NL$ are calculated using the parameters of the DNN, and $PI$ and $NI$ are the results of evaluating the similarity with the actual DNN parameter change values $ \Delta W_s $. These values represent the extent to which changes of the parameters cause ``changes in probability values that affect the inference result.'' Since $EF$ is calculated from the $PI$ and $NI$ of all input values, it represents the sum of the change in the probability values for each input value that affect the inference result.

Here, we note that input values in the same classification class are similar not only in the probability value but also in the amount of change in the probability value that flips the inference result from failure to success or vice versa. For example, for a dataset consisting of input values in the same classification class, the value of $EF$ that flips the inference result of an input value is assumed to be 5. Let us assume that the dataset consists of five input values and the value calculated as the sum of $EF$ values of those input values is 10. There can be three typical cases: $EF$ value of 2 from all 5 input values, $EF$ values of 10 from only 1 input value, and $EF$ values of 5 from 2 input values . In each case, the number of input values whose inference result flips is 0, 1, and 2, respectively. That is, when (the sum of) $EF$ is 10, it can be estimated that the number of input values whose inference result flips is from zero to two.

Next, consider the case where input values of different classification classes are mixed. In this case, the values of $EF$ that would flip the inference result is different for each input value. Suppose that if the dataset consists of the five input values, the values of $EF$ that flip the inference result is 1, 1, 3, 3, 10 for each input values, respectively. It is also supposed that the value calculated as the sum of $EF$ values of those input values is 10. In this case, the number of input values whose inference result flips could be any number from zero to four. In other words, in this case, we can only narrow down that the number of input values whose inference result flips to be any between zero and four when $EF$ is 10. Thus, when input values of different classification classes are mixed, it is more difficult to narrow down the number of input values for which the inference result flips from $EF$ compared with  the case where only input values of the same classification class are included. This means that the correlation between $EF$ and accuracy changes will be lower for a mixed dataset. For this reason, the proposed method is designed to be applied to each classification class.

We also realized from another experiment that applying the mini-batch method does not significantly change the $R^2$ score. This can also be explained by the fact that input values in the same classification class have similar features. If a mini-batch $X^{mb}$ was composed of various input values $x^m$ with different characteristics, positive losses $PL(x^m)$ were expected to be diverse. Since $PL(X^{mb})$ is calculated as the average of $PL(x^m)$, $PL(x^m)$ and $PL(X^{mb})$ are not similar for many $x^m$ in this case. However, mini-batches $X^{mb}$ consist of input values from the same classification class and their inference results are the same. Therefore, the $PL(x^m)$ calculated for each $x^m$ are all similar functions. Moreover, their average, $PL(X^{mb})$, is also similar to $PL(x^m)$. From these considerations, the value obtained by calculating $PI$ for each $x^m$ and summing them, and the value obtained by calculating $PI$ for each mini-batch $X^{mb}$ and multiplying it by the number of its elements $|X^{mb}|$ are highly likely to be similar. In other words, when the proposed method is applied to each classification class, the mini-batch method is likely to provide a similar result to the normal method. In fact, when we experimented with varying the number of input values that make up the mini-batch, no significant change in $R^2$ scores was observed.

$PL(x^m)$ is calculated for each input value $x^m$ included in the mini-batch $X_{F}^{mb}$ and the average value is used as $PL(X_{F}^{mb})$. Then, the gradient $ \nabla PL(X_{F}^{mb})$ is obtained for this average value. The gradient indicates how to update the parameters of the DNN so that it has a good (changing from failure to success) impact on average on the inference results of the input values contained in the mini-batch. It is used to compute the impact (i.e. $EF$) on the entire mini-batch by comparing it with the update of the DNN parameters. On the other hand, in the normal method, the gradient $ \nabla PL(x^m)$ indicates how to update the parameters so that it has a good impact on that input value. Recalling that the change in accuracy is the accumulation of the change in inference results for individual input values, the impact of the DNN update on individual input values is considered to have a higher correlation with the change in accuracy than the average impact on the mini-batch. For example, suppose that $\nabla PL(x^1)= \alpha $ and $\nabla PL(x^2)= -\alpha $ for a mini-batch consisting of $x^1$ and $x^2$. It is also assumed that the parameters of the DNN are updated in the direction that $PL(x^1)$ decreases. In that case, since $PL(x^2)$ increases, the inference result for $x^2$ remains a failure. On the other hand, since $PL(x^1)$ decreases, the inference result for $x^1$ is likely to change from failure to success, which may result in an increase in accuracy. However, if the mini-batch method is applied, since $ \nabla PL(X_{F}^{mb})= \alpha - \alpha = 0$, the value of $PL(X_{F}^{mb})$ does not change, that is, $PI(X_{F}^{mb}, N_s)=0$ even though $PL(x^1)$ decreases. Thus, the application of the mini-batch method can be a factor that reduces the correlation between $PI$ and changes in accuracy. The same is true with $NL$. Therefore, the $R^2$ score of the linear regression model is likely to be lower when the mini-batch method is applied.

In the calculations of $PL$ and $NL$, not only $t(x^m)$ but also $fst(y(x^m))$ and $snd(y(x^m))$ are used, respectively. This is because we expect to use cross entropy as loss function $L$. Cross entropy uses only the probability value of a particular classification class given as a supervisory signal. Therefore, for $PL$, calculating the loss on the basis of only $t(x^m)$ as the supervisory signal would only consider how much the value of $y_{t(x^m)}^m$ increases. This means that the loss does not include how much the value of $y_{fst(y(x^m))}^m$ decreases. Similarly, for $NL$, if only $t(x^m)$ is given as the supervisory signal and the loss is calculated on the basis of it, only by how much the value of $y_{t(x^m)}^m$ decreases is taken into account. The loss does not include by how much the value of $y_{snd(y(x^m))}^m$ increases. Therefore, the proposed method defines the calculations of $PL$ and $NL$ so that the values of $y_{t(x^m)}^m$, $y_{fst(y(x^m))}^m$, and $y_{snd(y(x^m))}^m$ are considered as factors affecting the change in accuracy. The mean squared error is calculated using probability values other than the class given as the supervisory signal. Therefore, if it is used as the loss function, it may be possible to remove terms in which $fst(y(x^m))$ and $snd(y(x^m))$ appear from the formulas of $PL$ and $NL$, respectively.

In the experiments shown in Section \ref{exp}, the original dataset was randomly split, so the distribution of the data does not change during incremental learning. This means that the effectiveness of the proposed method for concept drift was not evaluated. However, since the objective of the incremental learning shown in Section \ref{additional} is to adapt the DNN to concept drift as quickly as possible, it is assumed that the DNN is updated at a higher frequency than the frequency at which concept drift occurs. Hence, in many cases, the DNN will be updated even though concept drift has not occurred. Evaluating the effectiveness of the proposed method even when concept drift occurs is a future task.

\section{Related Work}
\label{relwork}
To the author's knowledge, no research on a method for quickly evaluating incremental learning results has been published. However, there are several works on incremental learning that focus on the parameter of DNNs (weights and biases) and the gradient of a loss function.

Kirkpatrick et al. \cite{overcoming} focused on a decrease in accuracy in task-incremental learning \cite{multitask1}\cite{multitask2}. For example, when training for task 2 is carried out after training for task 1, the performance of the previously trained task (task 1) is catastrophically reduced. This is called catastrophic forgetting \cite{cf01} \cite{cf02} \cite{cf03} \cite{cf04} \cite{aleixo2023catastrophic}. In response to this problem, they proposed a method of identifying the parameters important in regard to task 1, and training task 2 in a manner that minimize changes to those parameters (important in regard to task 1). Our proposed method does not distinguish the parameters of the DNN. It may be possible to estimate the accuracy change more accurately by identifying the parameters that contribute to the accuracy change for the dataset and focusing on the changes in those parameters as in their method.

In task-incremental learning and class-incremental learning, distillation loss is used to mitigate catastrophic forgetting \cite{mnemonics} \cite{castro2018end} \cite{icarl} \cite{aljundi2017expert} \cite{wu2019large} \cite{hou2018lifelong} \cite{dloss} \cite{rannen2017encoder} \cite{unified} \cite{distillation} \cite{dhar2019learning} \cite{douillard2020podnet} \cite{unified}. Distillation loss represents the difference between the inference results before and after learning. The more similar the inference results, the smaller the distillation loss value. Learning to minimize distillation loss in addition to the normal loss enables inference results for the past dataset to be preserved. 
Kang et al. proposed a learning method that focuses on the gradient of the loss function \cite{distillation}. In this method, learning is performed so that ``the increase in the loss'' for the past dataset when the DNN is updated is minimized. To achieve this, they focus on the gradient of the loss function for the past dataset. On the basis of this gradient, the change in loss for the past dataset is approximated and the DNN is updated so that it is minimized. In our proposed method, $PL$ and $NL$ are defined for the past dataset and the impact of DNN update on the past dataset is estimated on the basis of their gradients. From a general point of view, our method and Kang et al.'s method are similar in that they focus on the gradient of the loss function for the past dataset and estimate the impact of the DNN update on the basis of the gradient. 
Unlike this method, however, our proposed method aims to estimate the change in accuracy for the past dataset. For this purpose, we propose $PL$ and $NL$, which is directly related to the change in accuracy, rather than the normal loss.

In another approach, Belouadah et al. focused on that the weights of a DNN before updating represents the past classes in class-incremental learning \cite{classifier}. Their method, which uses weight to prevent catastrophic forgetting, is effective for memoryless class-incremental learning where the past dataset cannot be stored entirely. Our proposed method also utilizes the change in weight before and after DNN update, which is similar to their approach. However, our method cannot be applied to memoryless class-incremental learning since we cannot evaluate the accuracy for a past dataset without it. Although it is possible to apply the proposed method using part of the past dataset, the performance of the accuracy estimation is expected to be lower in that case.


\section{Conclusion}
\label{conclusion}
We proposed a method for quickly evaluating the effect of an additional training for the past dataset in incremental learning. In the proposed method, the gradient of the parameter values for the past dataset is extracted by running the DNN before the additional training. After the training, a barometer of the effect on the accuracy with respect to the past dataset is calculated from the gradient and update differences of the parameter values. Finally, the proposed method estimates the change in accuracy by using a regression model created from the $EF$s and actual changes in accuracy in the past trainings. The computational complexity of the proposed method after the training depends on the number of DNN parameters, not on the amount of data in the past dataset. Therefore, even if the amount of data included in the past dataset is enormous, applying the proposed method makes it possible to evaluate the effects of trainings quickly. When a DNN is updated during operation, the proposed method enables a system operator to decide whether to use the updated DNN with consideration of the change in accuracy for the past dataset. The results of our experiments indicate the usefulness of the proposed method in terms of computation time and the coefficient of determination for the regression model used to estimate changes in accuracy. Even though the expected coefficient of determination could not be confirmed in a number of cases, using the proposed method to obtain reference information for selecting a tentative DNN is preferable until an accurate change in accuracy is confirmed by test execution?]. As for future work, the proposed method will be more elaborately evaluated using other datasets. In particular, the occurrence of concept drift should be simulated in the evaluation. Moreover, improving the means of creating $EF$ will help in the search for a more accurate evaluation of the change in accuracy. For example, distillation loss discussed in Section \ref{relwork} may be effective for the improvement.

\bibliographystyle{IEEEtran}
\bibliography{quick}

\end{document}